\documentclass{article} 
\usepackage{iclr2025_conference,times}

\usepackage{amsmath,amsfonts,bm}

\def\eqref#1{equation~\ref{#1}}

\def\1{\bm{1}}

\DeclareMathAlphabet{\mathsfit}{\encodingdefault}{\sfdefault}{m}{sl}
\SetMathAlphabet{\mathsfit}{bold}{\encodingdefault}{\sfdefault}{bx}{n}

\usepackage{hyperref}
\usepackage{url}
\usepackage[utf8]{inputenc} 
\usepackage[T1]{fontenc}    
\usepackage{hyperref}       
\usepackage{booktabs}       
\usepackage{amsfonts}       
\usepackage{nicefrac}       
\usepackage{microtype}      
\usepackage{hyperref}
\usepackage{xcolor}         

\usepackage{colortbl}
\usepackage{wrapfig,multirow}
\usepackage{float}
\usepackage{graphicx}
\usepackage{enumitem}
\usepackage{booktabs}
\usepackage{tabularx}

\title{MIA-Bench: Towards Better Instruction \\ Following Evaluation of Multimodal LLMs}

\author{%
 Yusu Qian$^1$, Hanrong Ye$^{1,2}$, Jean-Philippe Fauconnier$^1$, \\ \textbf{Peter Grasch}$^1$, \textbf{Yinfei Yang}$^1$,
 \textbf{Zhe Gan}$^{1}$\\
$^1$Apple \quad $^2$HKUST\\
\texttt{\small \{yqian22,hanrong\_ye,jfauconnier,pgrasch,yinfei\_yang,z\_gan\}@apple.com} 
 }

\iclrfinalcopy 
\begin{document}

\maketitle

\begin{figure*}[h]
\centering
\includegraphics[width=\textwidth]{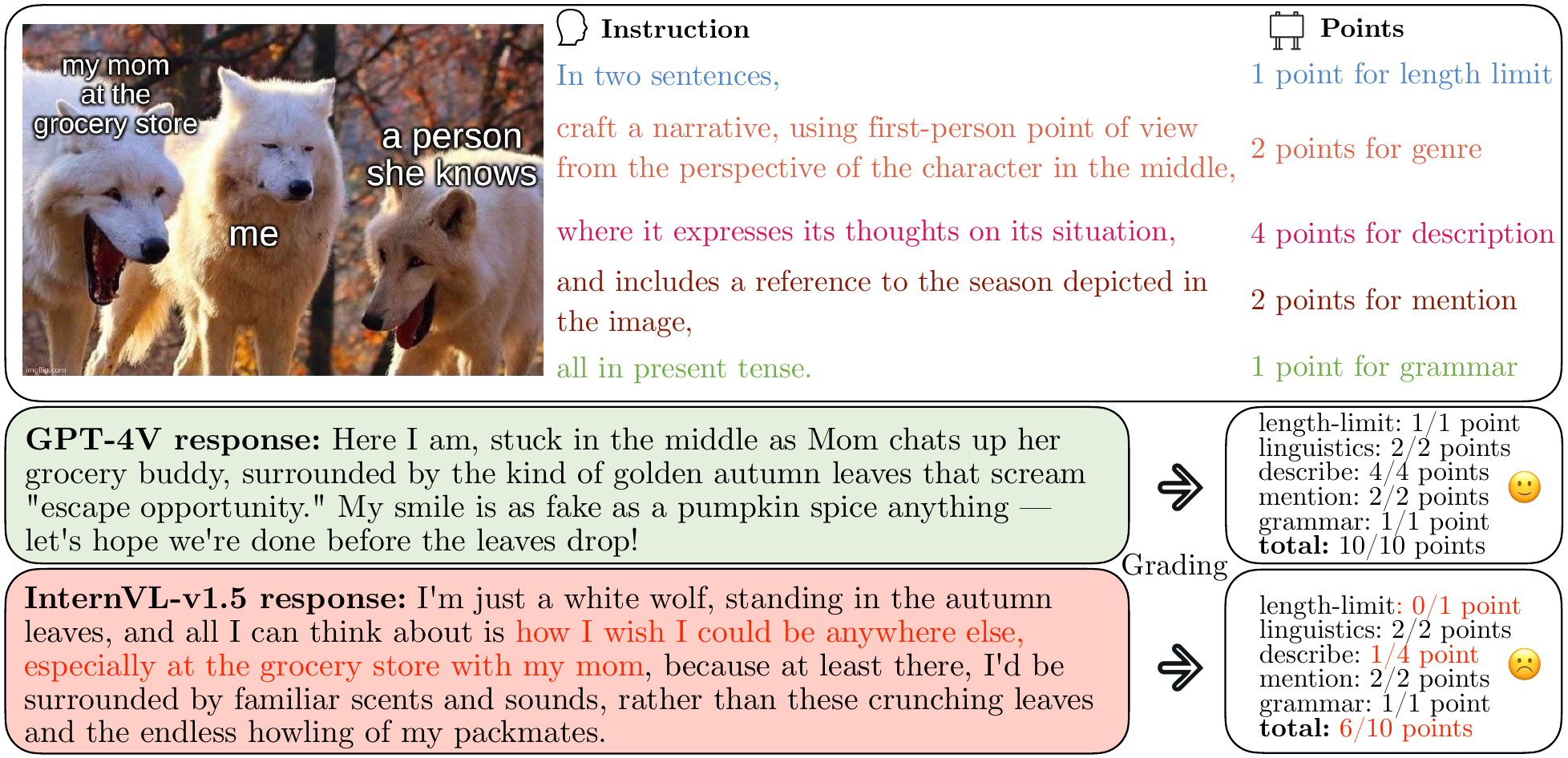}
\vspace{-0.3cm}
\caption{An example from MIA-Bench, featuring an image and a complex instruction to test models' compliance with layered instructions that are compositional in nature. Responses from GPT-4v~\citep{openai2024gpt4} and InternVL-v1.5~\citep{chen2024far} are evaluated using GPT-4o as the judge.} 
\label{fig:mia-bench-example}
\end{figure*}

\begin{abstract}
We introduce MIA-Bench, a new benchmark designed to evaluate multimodal large language models (MLLMs) on their ability to strictly adhere to complex instructions. Our benchmark comprises a diverse set of 400 image-prompt pairs, each crafted to challenge the models' compliance with layered instructions in generating accurate responses that satisfy specific requested patterns. Evaluation results from a wide array of state-of-the-art MLLMs reveal significant variations in performance, highlighting areas for improvement in instruction fidelity. Additionally, we create extra training data and explore supervised fine-tuning to enhance the models' ability to strictly follow instructions without compromising performance on other tasks. We hope this benchmark not only serves as a tool for measuring MLLM adherence to instructions, but also guides future developments in MLLM training methods.\footnote{Benchmark data and evaluation code: \url{https://github.com/apple/ml-mia-bench}.}
\end{abstract}
\section{Introduction}

\begin{figure*}[t!]
\centering
\includegraphics[width=\textwidth]{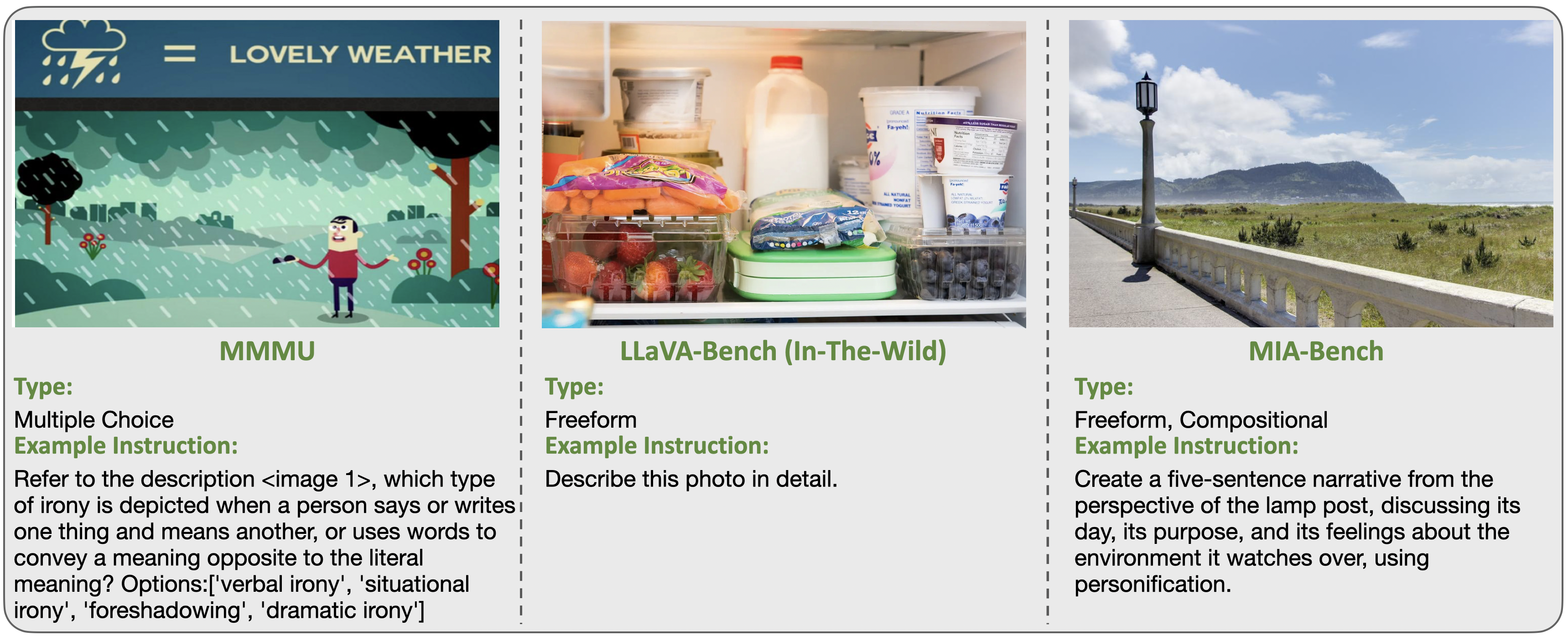}
\vspace{-20pt}
\caption{Comparison of various multimodal LLM benchmarks. (Left) Fixed-form visual question answering, often features short answers or multi-choice formats, such as MMMU~\citep{yue2023mmmu}. This format is popular due to its ease of evaluation. (Middle) Open-ended responses, such as LLaVA-Bench (in the Wild)~\citep{liu2023llava}. (Right) The proposed MIA-Bench, which also uses open-ended responses but focuses on evaluating precise adherence to complex instructions within the prompt.}
\label{fig:comparison_with_mmmu_and_llava_itw}
\vspace{-10pt}
\end{figure*}

The rapid advancement of Multimodal Large Language Models (MLLMs)~\citep{openai2024gpt4,liu2023llava,dai2023instructblip,liu2023improved,bai2023qwenvl,wang2023cogvlm,lin2023sphinx,geminiteam2023gemini,mckinzie2024mm1} has been a defining feature of recent AI research, showcasing increased model capabilities to comprehend and respond to visual inputs, often termed as multimodal ``instruction following''. 

To measure the progress of instruction following, many multimodal benchmarks have been developed, which can be roughly divided into two broad categories: ($i$) fixed-form visual question answering (VQA), often with short answers or using a multi-choice QA format; and ($ii$) free-form conversations with open-ended responses. Many current benchmarks have adopted the first format, including VQAv2~\citep{goyal2017making}, TextVQA~\citep{singh2019towards}, ScienceQA~\citep{lu2022learn}, MME~\citep{fu2023mme}, MMBench~\citep{liu2023mmbench}, SEED-Bench~\citep{li2023seed}, MathVista~\citep{lu2023mathvista}, and MMMU~\citep{yue2023mmmu}. These benchmarks are popular due to their ease of use in evaluating metrics and presenting model comparisons. 

However, as visual assistant models, the ability to engage users in free-form conversations is also crucial. Benchmarks in this format include LLaVA-Bench~\citep{liu2023llava}, MM-Vet~\citep{yu2023mm}, VisIT-Bench~\citep{bitton2023visitbench}, InfiMM-Eval~\citep{han2023infimm}, and the most recent Vibe-Eval~\citep{padlewski2024vibe} and LLaVA-Bench-Wilder~\citep{li2024llavanextstrong}. Typically, the free-form model responses are evaluated using external models as the judge.
These benchmarks are closer to daily-life visual chat scenarios; however, the type of ``instruction following'' examined in these benchmarks usually gauges a model's ability to perform tasks in a broad, often loosely defined manner. Yet, the precise adherence to complex instructions within prompts -- a critical aspect for evaluating LLMs~\citep{chia2023instructeval,zhou2023instructionfollowing,qin2024infobench} -- remains less explored in the context of multimodal LLMs. 

To this end, we introduce MIA-Bench,\footnote{Abbreviation for \textbf{M}ultimodal \textbf{I}nstruction \textbf{A}dherence Benchmark.}
 a new benchmark specifically designed for evaluating strict ``instruction adherence''.
Our instruction adherence metric measures the precision with which MLLMs can execute layered and compositional instructions.
This involves not only recognizing the content of the instructions, but also meticulously executing the detailed demands without deviation (\emph{e.g.}, answering in a given number of sentences, including specific elements, etc.). 
By establishing this stricter criterion, our benchmark aims to push the boundaries of model precision and reliability in practical applications, ensuring that outputs not only align with the general intent of the instructions, but also match the exact specifications provided. An example from MIA-Bench is provided in Figure \ref{fig:mia-bench-example}, and its comparison with previous MLLM benchmarks is illustrated in Figure \ref{fig:comparison_with_mmmu_and_llava_itw}.

MIA-Bench consists of 400 meticulously created image-prompt pairs, and encompasses diverse image contents including animals, food, landmarks, sport, art, landscape, text, etc. 
to cover a broad spectrum of real-world scenarios. In constructing this benchmark, we sought not only to evaluate the current capabilities of state-of-the-art MLLMs, but also to push the boundaries of what these models can achieve when rigorously tested against structured and layered instructions. The final prompts are of various complexity levels, and compositional in nature, with five base instruction categories, which are tailored to probe the models’ linguistic dexterity, grammatical accuracy, and descriptive fidelity. For example, the prompt in Figure \ref{fig:mia-bench-example} is composed of five base categories, including \textit{description}, \textit{mention}, \textit{grammar}, \textit{length limit}, and \textit{genre}.

We evaluate a wide array of MLLMs on the proposed benchmark, ranging from closed-source models (\emph{e.g.}, GPT-4o~\citep{gpt4o}, Gemini Pro~\citep{geminiteam2023gemini}, Claude-3~\citep{claude2022}, Reka~\citep{rekateam2024reka}) to open-source ones (\emph{e.g.}, LLaVA-NeXT~\citep{liu2024llavanext}, Intern-VL-Chat-1.5~\citep{chen2024far}, CogVLM2~\citep{wang2023cogvlm}, Phi-3-Vision~\citep{abdin2024phi3}). Our investigations reveal notable variations in model performance, highlighting great opportunities for improvement. 

To address these challenges, we further propose to generate training data tailored for supervised fine-tuning (SFT), where we aim to refine the models' abilities to process and comply with multifaceted instructions. Results from our SFT experiments indicate a promising enhancement in the models’ performance to strictly adhere to instructions, without hurting performance on other benchmarks. 

Our contributions are summarized as follows. ($i$) We construct MIA-Bench, a new benchmark to comprehensively evaluate MLLMs on their capability to strictly adhere to instructions. ($ii$) We provide a detailed analysis of popular MLLMs, and suggest training methods for enhanced instruction following. For this purpose, we created training data and conducted experiments for additional supervised fine-tuning. We hope this benchmark can serve as a useful
resource to stimulate further research on multimodal instruction adherence.
\section{MIA-Bench}

\begin{figure*}[t!]
\centering
\includegraphics[width=\textwidth]{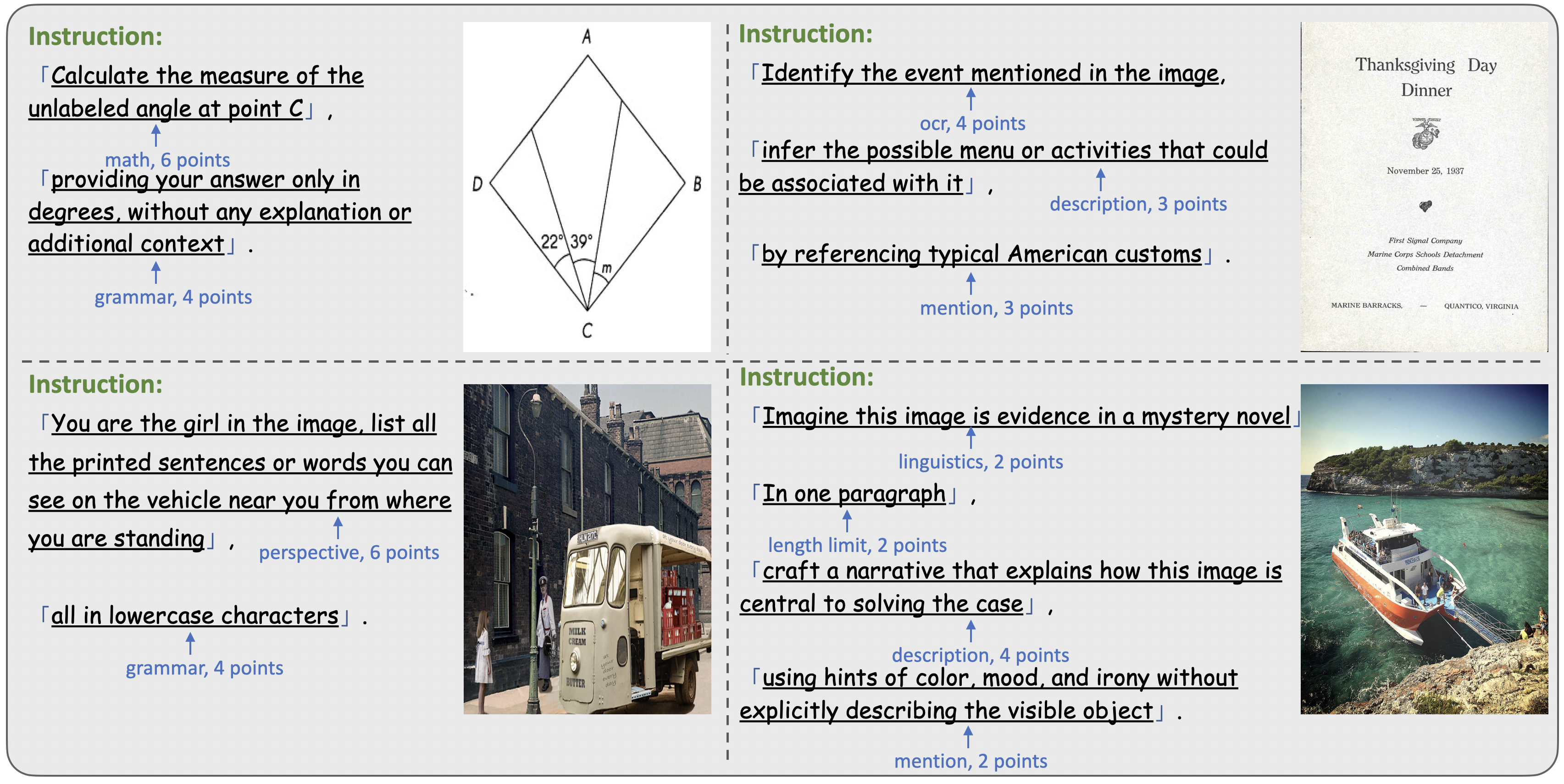}
\vspace{-20pt}
\caption{Examples from MIA-Bench, with detailed information on the instruction composition, base instruction weight and type. }
\label{fig:instruction_example_5}
\vspace{-10pt}
\end{figure*}

\begin{figure}[t!]
\centering
\vspace{-20pt}
\includegraphics[width=\textwidth]{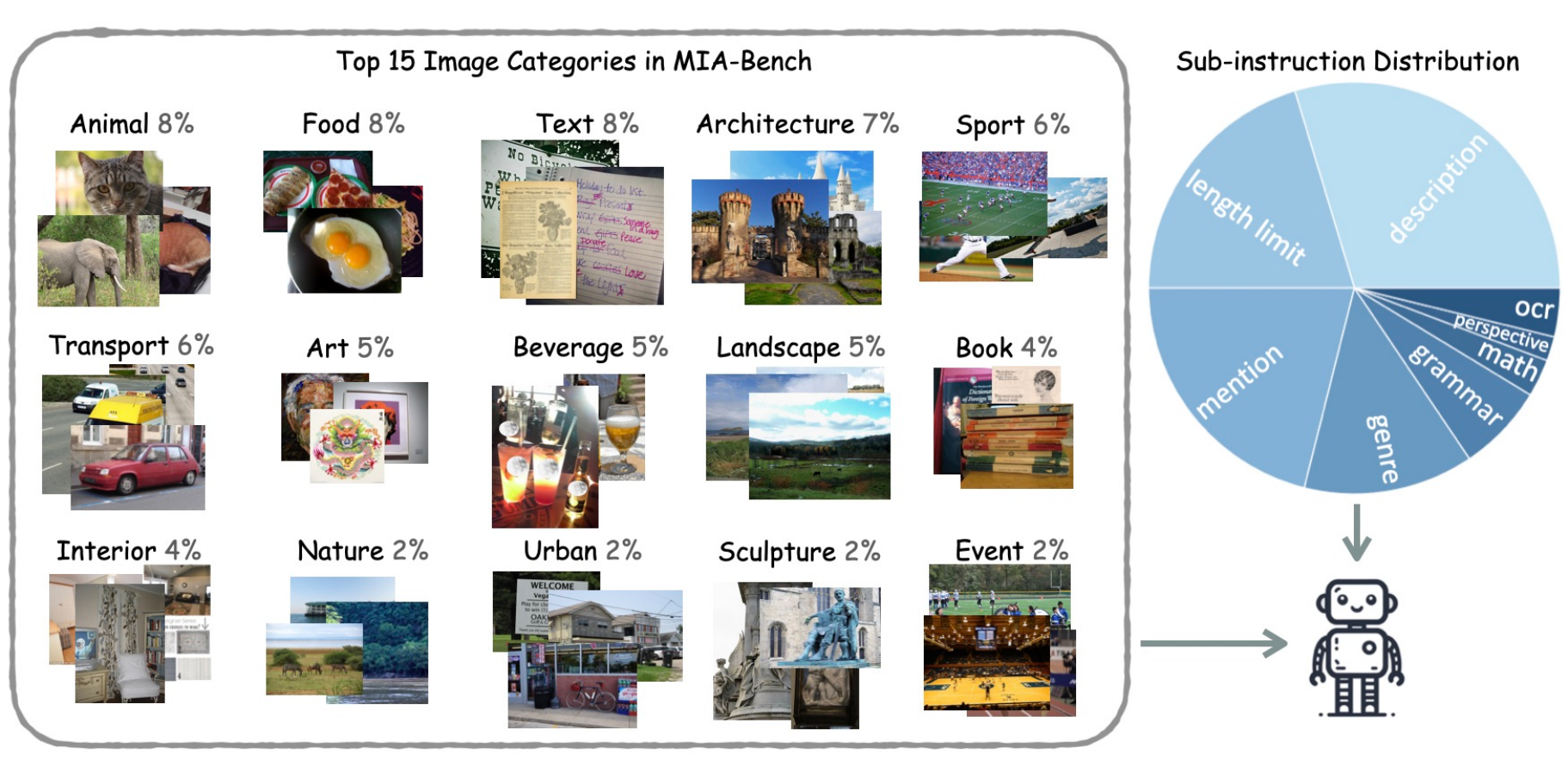}
\vspace{-20pt}
\caption{The top-15 image content categories and the distribution of the 8 sub-instruction categories in MIA-Bench.
}
\vspace{-10pt}
\label{fig:imagecategory}
\end{figure}

MIA-Bench consists of 400 image-prompt pairs, with examples shown in Figure \ref{fig:instruction_example_5}. The images are collected from diverse sources, including COCO 2017 validation set~\citep{lin2015microsoft}, SBU~\citep{Ordonez:2011:im2text}, TextVQA~\citep{Singh_2019_CVPR}, and Flickr. Images in the Flickr subset are photos of a variety of themes, including animals, art, architectures, text, food, math, etc. Images from the other three sources are randomly sampled from each corresponding source. 
Figure \ref{fig:imagecategory} shows the top 15 image content categories and the distribution of the 8 sub-instruction categories in MIA-Bench. The image content is labeled using GPT-4v. For each image, we manually write diverse and challenging instructions that contain multiple sub-instructions. 

When constructing the instructions, we follow three principles, detailed below. 
\begin{itemize}[leftmargin=*]
\item  \textbf{Correctness.} The instruction needs to be answerable by humans. For example, asking about objects that do not exist in the image makes the prompt unanswerable.
\item  \textbf{No answer leakage.} The instruction should not contain the answer to itself. \textit{`What color is the green object?'} is an example of answer leakage.
\item  \textbf{Image-dependent.} MMStar~\citep{chen2024right} pointed out that on some multimodal benchmarks, MLLMs can generate correct answers without accessing images half of the time. Multi-modal capabilities are necessary to correctly answer MIA-Bench prompts.
\end{itemize}

\subsection{Instruction Categories}

In this paper, we use \textit{instruction} to refer to the entire textual input, which in MIA-Bench can generally be viewed as a composition of multiple individual requests or constraints. We refer to these individual components as \textit{sub-instruction}. Instructions in MIA-Bench are of diverse complexity, and sub-instructions contained are of multiple categories, summarized in Figure \ref{fig:imagecategory}. 

The sub-instructions in MIA-Bench can be categorized into \textit{description}, \textit{length limit}, \textit{mention}, \textit{genre}, \textit{grammar}, \textit{math}, \textit{perspective}, and \textit{OCR}, detailed below.
\begin{itemize}[leftmargin=*]
\item \textit{`description'} refers to describing a certain part of the image, with the exception of text-rich parts of the image, which falls under the \textit{`OCR'};
\item \textit{`length limit'} refers to the limitation of response length (\emph{e.g.,} in exactly two sentences, using exactly 60 words); 
\item \textit{`mention'} refers to mentioning or not mentioning certain objects or entities (\emph{e.g.}, highlighting two similarities and one difference, comparing and contrasting the condition of the buildings with the activity on the street);
\item \textit{`genre'} refers to requests for a specific written form (\emph{e.g.}, write a poem, write a narrative, with at least one pun included, all while weaving in a subtle theme of change); 
\item \textit{`grammar'} refers to grammatical requirements (\emph{e.g.}, use present tense, use capitalized letters, use integers);
\item \textit{`math'} refers to requirements to come up with a solution to math problems, or to identify errors in solutions to math problems, or to generate a valid math problem given table, charts, etc.;
\item \textit{`perspective'} refers to requirements specifying the viewpoint of an object or person in the image. This requires MLLMs to correctly identify what can or cannot be seen from the specified position, and understand the spatial relationship of objects in its surrounding with itself (\emph{e.g.}, imagine you are the lady in the image, describe what you can see without turning your head around);
\item \textit{`OCR'} refers to requirements related to understanding OCR information in text-rich images such as menus, tickets, bills, etc. For example, given a photo of a ticket, the sub-instruction asking about the price printed on the ticket falls into this category. 
\end{itemize}

\begin{wrapfigure}{r}{0.45\textwidth}
\centering
\includegraphics[width=0.45\textwidth]{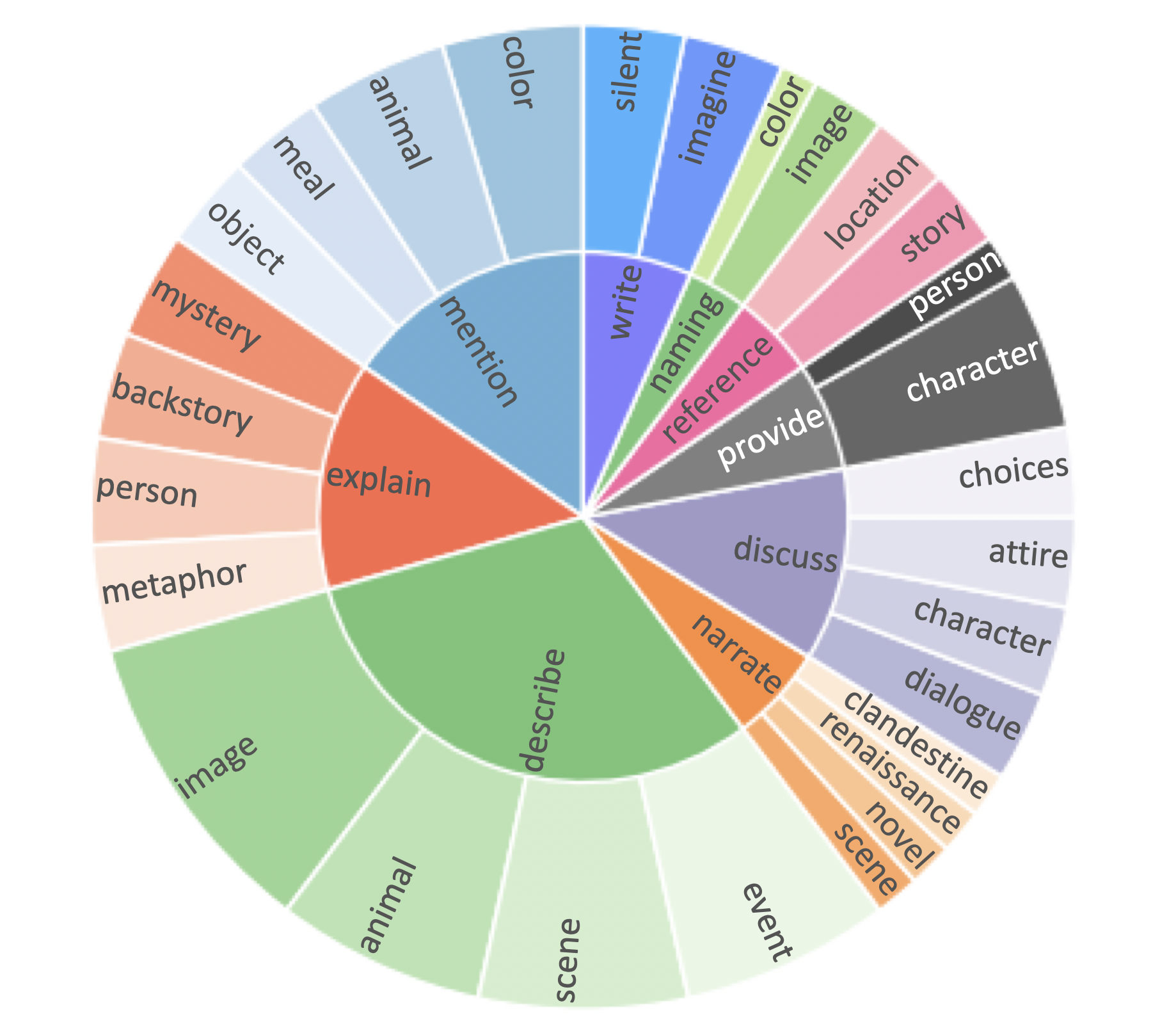}
\vspace{-20pt}
\caption{The most frequently used verbs and co-occurring nouns in MIA-Bench.}
\label{fig:word}
\end{wrapfigure}

Figure \ref{fig:word} shows the most frequently used verbs and co-occurring nouns in MIA-Bench.
To guarantee the diversity of prompts, when writing the instructions, we construct instructions of various levels of complexity: \textit{basic}, \textit{intermediate}, \textit{advanced}, \textit{creative}, and \textit{complex}. 
The \textit{basic} category is the simplest; the instructions normally only contain one or two sub-instructions, such as ``What is the color of the cat?'', or ``Describe the sofa in two words.''. The \textit{intermediate} category consists of instructions that contain three or more sub-instructions, but are in general easy for MLLMs to follow. 
The \textit{advanced} category contains instructions that are challenging and contain three or more sub-instructions. The \textit{creative} category contains instructions that instruct MLLMs to generate creative pieces of text, such as poems. 
The \textit{complex} category is a combination of the previous two categories; the instructions in this category are the most complicated as they usually contain multiple challenging sub-instructions. While we found these categories useful to elicit a diverse instruction set, we also found that practical examples were often difficult to categorize objectively. As a result, we only used these categories for data collection, but are not reporting per-category results.

\subsection{Response Evaluation Method}
To automatically evaluate MLLMs' performance with the proposed MIA-Bench at scale,
we adopt GPT-4o~\citep{gpt4o} as a judge model to score MLLMs' responses on each instruction and return a total score based on different criteria mentioned above.
Specifically, we design the following text prompt template:
\begin{figure}[H]
\centering
\vspace{-10pt}
\includegraphics[width=\textwidth]{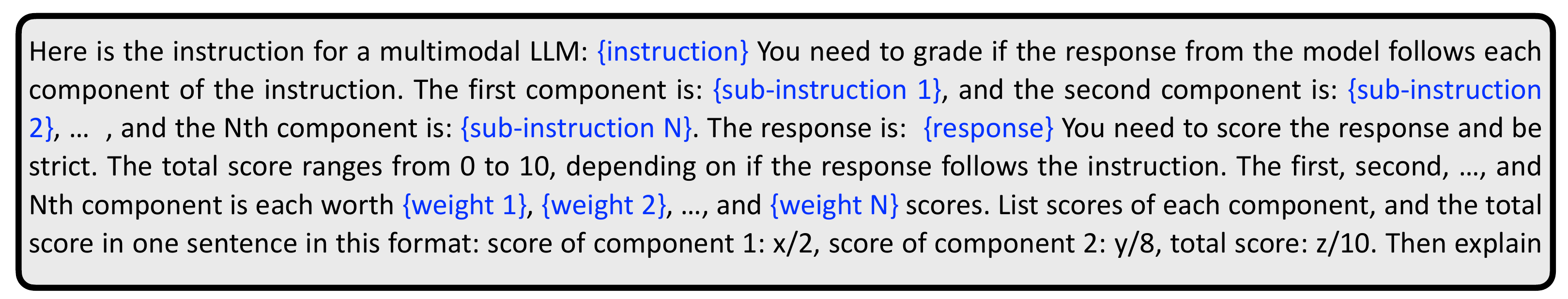}
\vspace{-26pt}
\label{fig:scoring_instruction}
\end{figure}
Each response is graded through a systematic process that begins by thoroughly evaluating how well it adheres to each specific sub-instruction provided within the overall task. This initial assessment focuses on the degree of compliance and alignment with the requirements outlined in the sub-instructions. After this detailed evaluation of individual components, a total score is computed by aggregating the performance across all sub-instructions.
Each sub-instruction within the broader instruction is assigned a maximum possible score, which can range from 1 to 10. The specific score assigned to a sub-instruction reflects its relative significance and difficulty within the context of the overall task. To ensure fairness and consistency, the total weight of all sub-instructions for any given instruction is set to sum up to 10.
This carefully calibrated scoring system has been meticulously designed to capture and reflect the varying levels of complexity and importance associated with each aspect of the task. By doing so, it ensures that more critical or challenging elements are appropriately weighted and that the evaluation process provides a comprehensive and accurate representation of the response's quality.

\begin{figure}[t!]
\centering
\vspace{-30pt}
\includegraphics[width=\textwidth]{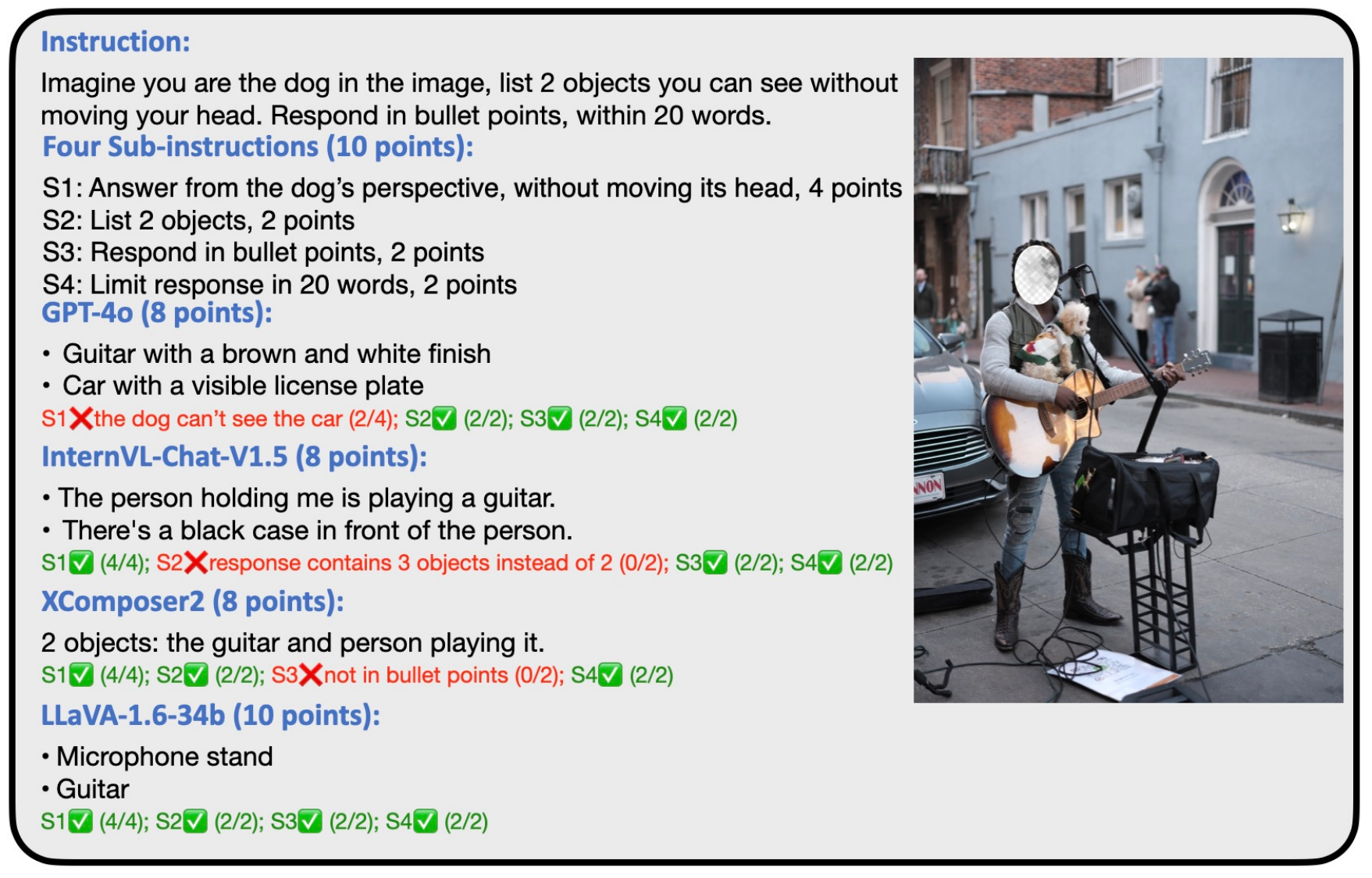}
\vspace{-20pt}
\caption{An example featuring responses from four MLLMs along with their evaluation scores, presented in a point-by-point format.
}
\vspace{-10pt}
\label{fig:visual_example}
\end{figure}

Figure \ref{fig:visual_example} shows an example of how responses from different MLLMs are evaluated and scored.
For example, length limits are often binary in nature—either met or not met—hence the single point allocation. In contrast, a description task may require the model to handle multiple layers of complexity, including accuracy, detail, and relevance, which justified a higher score. For the example in Figure \ref{fig:visual_example}, there are 4 sub-instructions (denoted from S1 to S4); the first is worth 4 points and the rest is worth 2 points each. The response from GPT-4o partially follows the first sub-instruction which requires the response to be from the perspective of the dog, as the dog should not be able to see the car behind the man without turning around. The dog should be able to see the guitar, thus GPT-4o gets 2 points out of 4 for the first sub-instruction. It successfully follows the other 3 sub-instructions, achieving full score for them. Thus, the final score GPT-4o reaches is 8 out of 10. We always assign larger weight (6 if there are two sub-instructions, 4 if there are three or more sub-instructions) to the sub-instruction in the \textit{description} category unless this category is absent in some cases, as usually a major part of the response is addressing this sub-instruction. For each MLLM, we compute the average score it gets on all 400 responses, and represent the ratio of the average score divided by 10 using percentage. We also compute the average score for each instruction category.

\section{Experiments}
In this section, we first present results of different MLLMs on MIA-Bench in Section~\ref{sec:results}, with additional supervised fine-tuning exploration in Section~\ref{sec:sft}.

\subsection{Benchmark Results}\label{sec:results}
\begin{table}[h]
\centering
\vspace{-30pt}
\huge
\begingroup
\renewcommand{\arraystretch}{1.7} 
\resizebox{\columnwidth}{!}{%
\begin{tabular}{cccccccccc}
\hline
  \textbf{Model} &
  \multicolumn{1}{c}{\textbf{Meta-Avg}} &
  \multicolumn{1}{c}{\textbf{Description}} &
  \multicolumn{1}{c}{\textbf{Len-Limit}} &
  \multicolumn{1}{c}{\textbf{Genre}} &
  \multicolumn{1}{c}{\textbf{Grammar}} &
  \multicolumn{1}{c}{\textbf{Mention}} &
  \multicolumn{1}{c}{\textbf{Math}} &
  \multicolumn{1}{c}{\textbf{Perspective}}&
  \multicolumn{1}{c}{\textbf{OCR}}\\ 
\rowcolor[HTML]{C0C0C0}\multicolumn{10}{c}{Open Source 1b\texttt{-}8b}\\ 
Fuyu-8b~\citep{fuyu-8b}    & 24.52&	52.06&	24.52&	17.06	&17.18	&36.43&	22.62	&66.67&	33.09
 \\ 
\rowcolor[HTML]{EFEFEF}Kosmos-2~\citep{peng2023kosmos}        & 26.06&	50.95&	38.52&	11.55	&19.78&	28.70	&17.26&	50.83	&41.88\\ 
Sphinx~\citep{lin2023sphinx}       & 50.99&	75.33&	53.51	&60.45&	48.28&	57.75&	47.41&	70.00&	61.04\\ 
\rowcolor[HTML]{EFEFEF}Idefics-2-8b~\citep{laurençon2024matters}     & 51.42	&59.37&	62.73&	48.07&	64.09&	46.20&	46.51&	48.33&	61.97 \\ 
mPLUG-Owl2~\citep{ye2023mplug}       & 57.86&	75.01&	65.25&	63.39&	60.26&	57.70&	57.22&	65.00&	62.08\\ 
\rowcolor[HTML]{EFEFEF}CogVLM-Chat~\citep{wang2023cogvlm} &  58.95	&60.42	&57.86&	67.94&	60.55&	62.92&	36.67&	60.83&	61.87 \\
ShareGPT4V~\citep{chen2023sharegpt4v} & 59.41&	81.08&	63.49&	63.88&	58.46&	62.49&	52.98&	82.50&	72.29 \\
\rowcolor[HTML]{EFEFEF}DeepSeek-VL-7b-chat~\citep{lu2024deepseekvl} & 60.96&	86.31&	63.26&	72.11&	54.79&	63.75&	67.39&	74.17&	77.85 \\
LLaVA-1.5-7b~\citep{liu2023llava}         & 62.18&	78.00&	68.60	&63.95&	64.18&	65.89&	47.31&	86.67&	60.75 \\
\rowcolor[HTML]{EFEFEF}LLaVA-NeXT-7b-vicuna~\citep{liu2024llavanext}  & 62.27&	79.21	&68.01	&65.63&	60.95	&63.33	&46.67&	90.00&	65.54 \\ 
Qwen-VL-Chat~\citep{bai2023qwenvl}       & 63.09	&80.51&	74.22&	66.95&	63.11&	63.01	&45.00&	75.83&	66.01\\ 
\rowcolor[HTML]{EFEFEF}XComposer2-7b~\citep{internlmxcomposer2_4khd} & 67.71&	83.47&	76.16&	73.66&	67.69	&67.01	&48.61&	77.50	&68.06 \\
CogVLM2~\citep{wang2023cogvlm} & 73.43&	87.60&	74.52	&83.47	&71.97&	77.01&	71.53	&90.83&	87.16 \\
\rowcolor[HTML]{EFEFEF}Phi-3-vision~\citep{abdin2024phi3} & 76.02	&84.90	&84.46	&86.52&	67.93&	74.70	&78.16&	74.17	&83.96 \\
MiniCPM-Llama3-v2.5~\citep{viscpm} & 76.27	&84.12	&79.44&	80.33	&81.25	&76.99	&64.08	&81.67&	76.59 \\

\rowcolor[HTML]{C0C0C0}\multicolumn{10}{c}{Open Source 8b\texttt{-}13b}\\ 
InstructBLIP-13b~\citep{dai2023instructblip}       & 38.16&	50.54	&39.57&	29.34	&38.43&	42.28&	12.50	&50.00	&30.42 \\ 
\rowcolor[HTML]{EFEFEF}LLaVA-1.5-13b~\citep{liu2023llava}         & 63.55&	80.98	&70.15&	64.54&	59.30&	67.42	&45.11	&69.17&	76.28 \\ 
LLaVA-NeXT-13b-vicuna~\citep{liu2024llavanext} & 69.16&	86.75&	69.88&	82.07	&64.77	&74.99	&48.56&	77.50&	75.83 \\ 

\rowcolor[HTML]{C0C0C0}\multicolumn{10}{c}{Open Source 13b\texttt{-}110b}\\ 

Yi-VL-34b~\citep{ai2024yi}&53.90&	74.89&	52.05&	59.09&	55.91&	57.25&	54.17&	41.85&	70.09 \\
\rowcolor[HTML]{EFEFEF}InternVL-Chat-v1.5~\citep{chen2024far}  & 75.42&	89.13&	78.21&	79.92	&78.16&	77.54&	76.11&	87.50&	80.92 \\ 
LLaVA-NeXT-34b~\citep{liu2024llavanext}        & 75.61&	88.02	&83.50	&86.58&	71.57	&75.83	&68.06	&87.50	&80.26 \\ 
\rowcolor[HTML]{EFEFEF}LLaVA-NeXT-110b~\citep{liu2024llavanext} & 79.84	&86.99	&84.86	&82.49&	79.04&	80.10&	71.94&	80.83	&75.45\\

\rowcolor[HTML]{C0C0C0}\multicolumn{10}{c}{Proprietary}\\ 
Gemini-1.0-Pro~\citep{geminiteam2023gemini}           & 70.63&	82.77&	72.83&	78.76&	76.91	&71.67&	81.45	&89.29&	84.11 \\ 
\rowcolor[HTML]{EFEFEF}Reka-Core~\citep{rekateam2024reka} & 76.95&	\textbf{91.05}	&79.91&	85.16&	78.98	&82.08&	82.53&	77.50	&81.08 \\
Claude-3-Haiku~\citep{claude2022}      & 78.25	&86.86	&77.53&	90.27&	73.41	&82.62&	82.22	&57.50&	86.49 \\ 
\rowcolor[HTML]{EFEFEF}Claude-3-Sonnet~\citep{claude2022}       & 79.44	&88.06	&82.71&	90.54	&79.60&	82.05&	82.22	&76.67&	84.43  \\ 
Claude-3-Opus~\citep{claude2022}     & 84.50&	90.50&	86.03&	91.19	&83.82&	85.49	&85.92	&65.00	&\textbf{86.84}\\ 
\rowcolor[HTML]{EFEFEF}GPT-4v~\citep{openai2024gpt4}  & 86.11	&90.03	&87.61&	\textbf{94.59}&	80.12	&89.37	&85.63	&59.17&	85.26 \\
GPT-4o~\citep{gpt4o} & \textbf{88.58}	&90.82&	\textbf{92.73}	&94.29&	\textbf{85.70}	&\textbf{90.66}	&\textbf{87.07}	&\textbf{92.50}	 & 86.54\\
\hline
\end{tabular}%
}
\endgroup
\vspace{-3mm}
\caption{Evaluation results of a wide array of MLLMs on MIA-Bench.}
\label{tab:mia-bench-results}
\vspace{-10pt}
\end{table}

In total, we have evaluated 29 popular MLLMs on MIA-Bench. Results are reported in Table~\ref{tab:mia-bench-results}. 
Observations are summarized as follows.
\begin{itemize}[ leftmargin=*]
\item Overall, the best performance was achieved by GPT-4o~\citep{gpt4o}, with a score 88.58, showcasing its superiority across different categories of instruction adherence. 
\item  The ability to describe content accurately was best exhibited by Reka~\citep{rekateam2024reka}. Other models like Claude-3-Opus~\citep{claude2022}, GPT-4v~\citep{openai2024gpt4} and GPT-4o also achieved scores higher than 90. This suggests that these models are good at generating coherent and contextually appropriate text. 
\item In the genre category, the highest proficiency was shown by GPT-4v and GPT-4o with scores above 94, suggesting an exceptional grasp of language nuances. Among open-source models, Phi-3-Vision~\citep{abdin2024phi3} and LLaVA-NeXT-34b~\citep{liu2024llavanext} show strong performance with scores of 86.52 and 86.58, respectively. The lowest score on this metric was by Kosmos-2~\citep{peng2023kosmos}, with a mere 11.55, pointing to difficulties in understanding or generating linguistically complex sentences.
\item GPT-4o excelled in grammar with a score of 85.70, which indicates superior ability in syntax correctness and sentence structuring that matches specific requirements in the instruction. Among the open-source models, MiniCPM-Llama3-V-2.5~\citep{viscpm} is notable with a score of 81.25. Contrastingly, Fuyu-8b~\citep{fuyu-8b} scored the lowest with 17.18, reflecting major challenges in grammar adherence.
\item GPT-4o also showed the best performance with a score of 92.73 in respecting prescribed length limits, which is crucial for tasks requiring concise and precise answers.  Among open-source models, LLaVA-NeXT-110b~\citep{liu2024llavanext} stands out with a score of 84.86.
\item Results from LLaVA series also suggest a strong correlation between LLM size and MIA-Bench performance across metrics. 
\end{itemize}

\begin{table}[]
\centering
\begingroup
\renewcommand{\arraystretch}{1.5}

\resizebox{\columnwidth}{!}{%
\begin{tabular}{cccccccccc}
\hline
\multirow{2}{*}{\textbf{Model}} & \multirow{2}{*}{\textbf{MME}} & \multirow{2}{*}{\textbf{MMMU}} & \multirow{2}{*}{\textbf{MMB}} & \multirow{2}{*}{\textbf{MMVet}} & \multirow{2}{*}{\textbf{HallB}} & \multirow{1}{*}{\textbf{Math}} & \multirow{1}{*}{\textbf{Meta}} & \multirow{2}{*}{\textbf{MIA}} & \multirow{1}{*}{\textbf{MIA}}\\
 &  &  & &  &  & \multirow{1}{*}{\textbf{Vista}} & \multirow{1}{*}{\textbf{Ranking}} & \multirow{2}{*} & \multirow{1}{*}{\textbf{Ranking}}
\\
\rowcolor[HTML]{EFEFEF}GPT-4o~\citep{gpt4o} & 2328.7&	69.1&	83.3/82.1&	66.5	&67.5	&63.8	&1	&88.58&	1
\\ 
GPT-4v~\citep{openai2024gpt4} & 1926.6	&56.8&	77/74.4&	67.6&	46.5	&49.9	&3	&86.11&	2
  \\ 
\rowcolor[HTML]{EFEFEF}Gemini-1.0-Pro~\citep{geminiteam2023gemini} & 1933.4&	47.9&	73.6/74.3&	64.3	&45.2	&45.2	&5*	&70.63&	6
\\ 
Claude-3-Opus~\citep{claude2022} & 1586.8	&59.4&	63.3/59.2&	58.1	&37.8	&50.5	&5*	&84.50&	3 \\
\rowcolor[HTML]{EFEFEF}InternVL-Chat-V1-5~\citep{chen2024far} & 2187.8	&45.2	&82.2/82&	62.8&	49.3&	53.5&	2	&75.42&	5 \\
LLaVA-NeXT-34b~\citep{li2024llavanextstrong} &2028&	51.1	&81.1/79	&48.9	&47.6	&47.7	&4	&75.61	&4  \\
\hline
\end{tabular}%
}
\endgroup
\vspace{-3mm}
\caption{Meta ranking of five state-of-the-art MLLMs on existing multimodal benchmarks compared with their ranking on MIA-Bench.}
\label{tab:benchmark_ranking}
\vspace{-10pt}
\end{table}

\textbf{Correlation with other benchmarks.}
In Table \ref{tab:benchmark_ranking}, we compare the ranking of 5 state-of-the-art MLLMs on MIA-Bench as well as their meta ranking on MME~\citep{fu2023mme}, MMMU~\citep{yue2023mmmu} , MMBench~\citep{liu2024mmbench}, MMVet~\citep{yu2023mm}, HallusionBench~\citep{guan2023hallusionbench}, and MathVista~\citep{lu2023mathvista} (meta ranking is computed by averaging rankings across these benchmarks). Our findings reveal a discrepancy between the two sets of rankings. Notably, InternVL-Chat-V1.5~\citep{chen2024far}, which holds the highest meta-ranking among the five MLLMs on the other benchmarks, ranks the lowest on MIA-Bench. Conversely, Claude-3-Opus, which has the lowest meta-ranking, secures the second position on the MIA-Bench. This indicates that excelling in tasks evaluated by existing benchmarks does not necessarily translate to superior instruction adherence capability assessed by MIA-Bench.

\begin{wrapfigure}{r}{0.5\textwidth}
\centering
\includegraphics[width=0.5\textwidth]{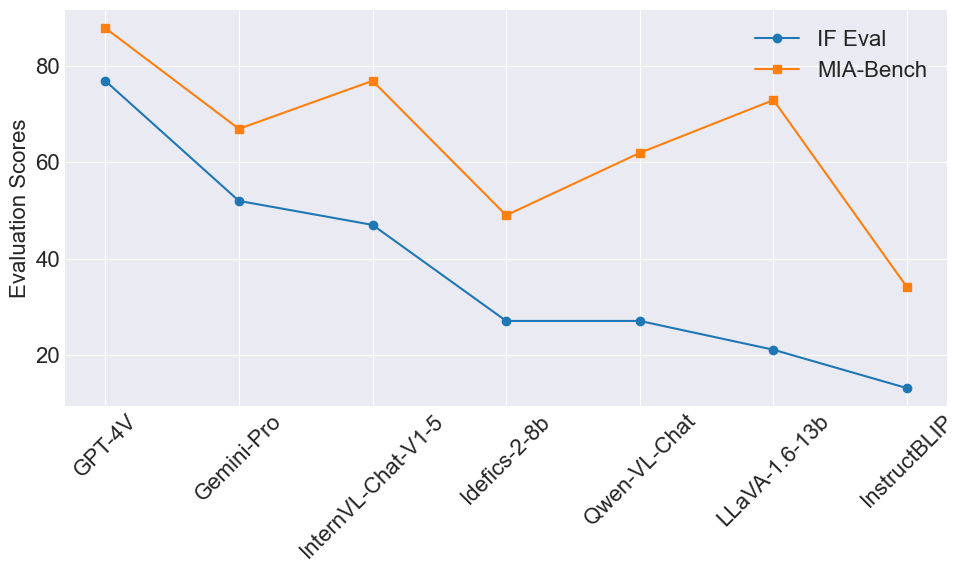}
\vspace{-20pt}
\caption{Study of correlation between scores on MIA-Bench and IFEval.}
\vspace{-10pt}
\label{fig:ifeval2}
\end{wrapfigure}
\textbf{Correlation with LLM backbone performance.}
To determine if the performance on MIA-Bench is attributable solely to the underlying LLMs, we also evaluate several MLLMs on IFEval~\citep{qin2024infobench}, a benchmark that assesses the instruction adherence capability of LLMs, and compare their ranking with that on MIA-Bench. This comparison is shown in Figure~\ref{fig:ifeval2}, which shows that the instruction adherence capabilities of MLLMs do not consistently align with their LLMs' adherence capability.

\textbf{Other external models as the judge.}
Since the evaluation uses GPT-4o as the judge, it is natural to conjecture that GPT-4o may favorably score its own responses. To alleviate this concern, we use Claude-3, a strong performer in Table \ref{tab:mia-bench-results}, to evaluate responses from GPT-4o and itself, and compare their scores with each other. The prompt used to grade responses is the same as the one used in GPT-4o grading. We find that even using Claude-3 Opus to score its own and GPT-4o's responses, GPT-4o still achieves a superior score. When scored by Claude-3-Opus, GPT-4o achieves 89.84 score in contrast to Claude-3-Opus' 85.89. Based on this observation, we use GPT-4o for evaluation by default, and observe that results from multiple runs may have around 1\% variation.

We further extend our study by evaluating the models using various versions of ChatGPT as judge models, including  gpt-4o-mini-2024-07-18, gpt-4o-2024-05-13, gpt-4o-2024-11-20, and chatgpt-4o-latest, and show the scores and ranking in Table~\ref{tab:4_gpt_version_total_scores} with detailed results in Table~\ref{tab:model_scores_gpt-4o-mini-2024-07-18},~\ref{tab:model_scores_gpt-4o-2024-05-13},~\ref{tab:model_scores_gpt-4o-2024-11-20} and \ref{tab:model_scores_chatgpt-4o-latest}. We find that the ranking is consistent even though the judge models are different.

\begin{table}[h]
\centering
\begingroup
\resizebox{\columnwidth}{!}{%
\begin{tabular}{ccccccccc}
\hline
\multirow{2}{*}{\textbf{Model}} & 
\multicolumn{2}{c}{\textbf{chatgpt-4o-latest}} & \multicolumn{2}{c}{\textbf{gpt-4o-2024-11-20}} & \multicolumn{2}{c}{\textbf{gpt-4o-2024-05-13}} & \multicolumn{2}{c}{\textbf{gpt-4o-mini-2024-07-18}} \\
\cline{2-9}
 & \textbf{Score} & \textbf{Ranking} & \textbf{Score} & \textbf{Ranking} & \textbf{Score} & \textbf{Ranking} & \textbf{Score} & \textbf{Ranking} \\
\rowcolor[HTML]{EFEFEF}GPT-4o & 89.69 & 1 & 89.94 & 1 & 90.97 & 1 & 81.36 & 1 \\

Claude-3-Opus & 86.16 & 2 & 84.89 & 2 & 85.61 & 2 & 78.95 & 2 \\

\rowcolor[HTML]{EFEFEF}Reka & 83.09 & 3 & 82.68 & 3 & 83.99 & 3 & 77.70 & 3 \\

MiniCPM-Llama3-V2.5 & 78.10 & 4 & 78.75 & 4 & 79.80 & 4 & 73.72 & 4 \\

\rowcolor[HTML]{EFEFEF}Gemini & 75.77 & 5 & 76.32 & 5 & 77.36 & 5 & 67.45 & 5 \\

LLaVA-1.5-13b & 66.78&	6&	66.05&	6&	68.39&	7	&61.54&	6 \\

\rowcolor[HTML]{EFEFEF}ShareGPT4v & 66.61 & 7 & 65.72 & 7 & 68.90 & 6 & 60.30 & 7 \\

Idefics-2-8b & 53.51 & 8 & 53.61 & 8 & 54.18 & 8 & 44.28 & 8 \\

\hline
\end{tabular}%
}
\endgroup
\vspace{-3mm}
\caption{Comparison of scores and rankings across different judge models. The ranking is stable.}
\label{tab:4_gpt_version_total_scores}
\end{table}

\subsection{Supervised Fine-Tuning (SFT)}\label{sec:sft}
The performance of small-scale models such as LLaVA-NeXT-13b is less ideal on MIA-Bench. In this section, we study the use of supervised fine-tuning to enhance model performance.

\textbf{Additional SFT data construction.}
First, we randomly sample 1000 images from COCO 2017 training set, and use GPT-4v to generate five instructions for each image, using the prompt below. 

\begin{table}[t!]
\vspace{-20pt}
\centering
\huge
\renewcommand{\arraystretch}{1.5} 
\resizebox{\columnwidth}{!}{%
\begin{tabular}{lccccccccc}
\hline
  \textbf{Model} & \textbf{Total Score} & \textbf{Description} & \textbf{Length Limit} & \textbf{Genres} & \textbf{Grammar} & \textbf{Mention} & \textbf{Math} & \textbf{Perspective} & \textbf{OCR}\\ %
\rowcolor[HTML]{EFEFEF}LLaVA-NeXT-13b~\citep{liu2024llavanext} & 69.16&	86.75&	69.88&	82.07	&64.77	&74.99	&48.56&	77.50&	75.83 \\ 
~~~~+ DIT & 78.85 & 86.90 & 86.80  & 88.02  & 71.34 & 81.01 & 60.87 & 84.17 & 72.65 \\ 
\rowcolor[HTML]{EFEFEF}~~~~+ DIT + LLaVA-Instruct150k & 78.90 & 88.59 & 74.67  & 79.95  & 74.17 & 66.39 & 53.70 & 100.00 & 80.83 \\ \hline
\end{tabular}%
}
\vspace{-3mm}
\caption{Detailed results of LLaVA-NeXT-13b~\citep{liu2024llavanext} on MIA-Bench before and after supervised fine-tuning on additional constructed diverse instruction-tuning (DIT) data, and the mixture of DIT and LLaVA-Instruct150k. We re-ran the baseline.}
\label{tab:Evaluation result of MLLMs before and after SFT}
\end{table}

\begin{table}[t!]
\centering
\renewcommand{\arraystretch}{1.5} 
\resizebox{\columnwidth}{!}{%
\begin{tabular}{lcccccccccc}
\hline
  \textbf{Model} & \textbf{MMBench} & \textbf{TextVQA} & \textbf{VQA2} & \textbf{LLaVA-itw} & \textbf{POPE} & \textbf{VizWiz}  & \textbf{MIA-Bench} \\ %
\rowcolor[HTML]{EFEFEF}LLaVA-NeXT-13b~\citep{liu2024llavanext} & 70.6 &	64.26&	82.80 &	85.8  & 87.7	&60.41	 &69.16\\ 
~~~~+ DIT& 68.6 & 63.20 & 82.58  & 83.4  & 86.9 & 59.72  & 78.85 \\ 
\rowcolor[HTML]{EFEFEF}~~~~+ DIT + LLaVA-Instruct150k & 67.27 & 54.24 & 77.92  & 75.8  & 87.8 & 58.87  & 78.90 \\ \hline
\end{tabular}%
}
\vspace{-3mm}
\caption{Results of LLaVA-NeXT-13b~\citep{liu2024llavanext}  on MIA-Bench and other major multimodal benchmarks supervised fine-tuning on additional constructed diverse instruction-tuning (DIT) data, and the mixture of DIT and LLaVA-Instruct150k. We re-ran the baseline.}
\label{tab:my-table-SFT}
\end{table}

\begin{figure}[H]
\vspace{-3mm}
\centering
\includegraphics[width=\textwidth]{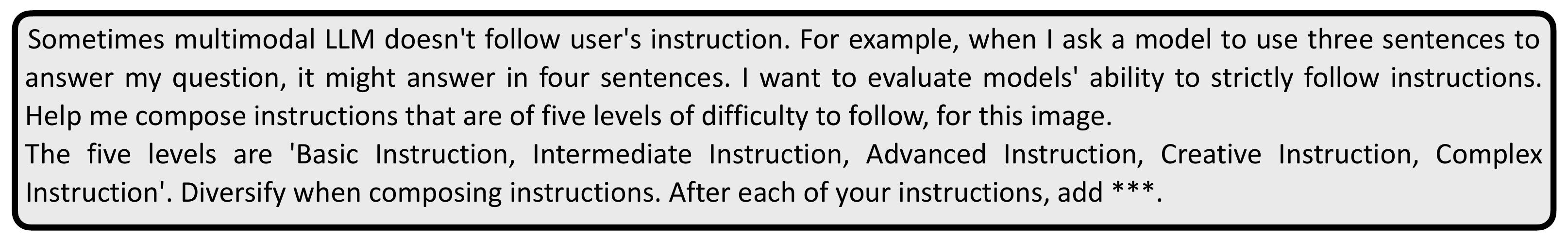}
\end{figure}
\vspace{-5mm}
We then manually process the generated instructions. The cleaned data for SFT consists of 5000 image-prompt pairs.

Then, we use GPT-4v to generate responses to the constructed prompts. To evaluate the quality of these responses, we sampled 100 responses and manually checked if they adhere to the instructions. We find that 90\% of the responses successfully followed all instructions in the prompt, serving as a proper ground-truth response for model training. Examples of this additional training data is provided in the Appendix.

\begin{figure*}[t!]
\centering
\includegraphics[width=\textwidth]{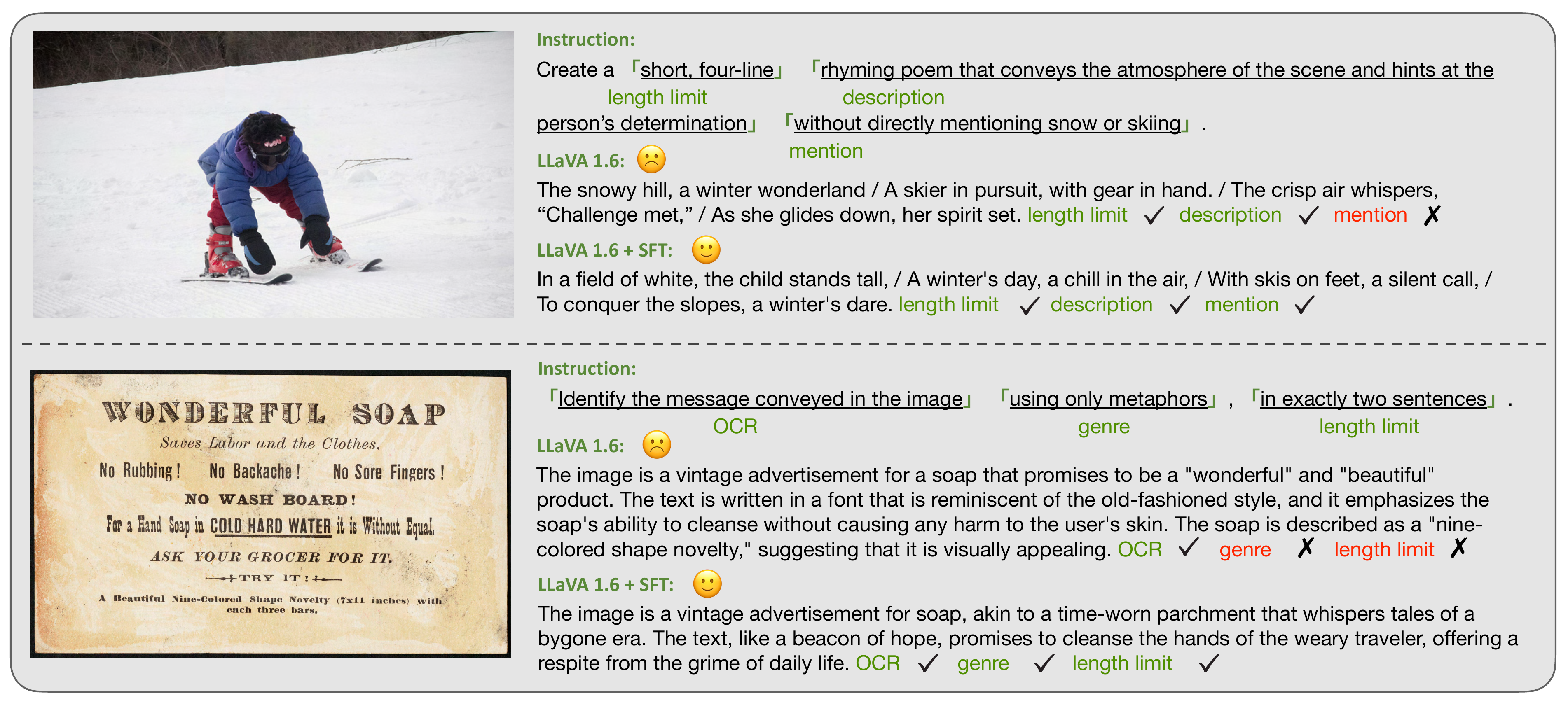}
\vspace{-20pt}
\caption{Examples of LLaVA-NeXT-13b responses before and after supervised fine-tuning on additional diverse instruction-tuning data. }
\vspace{-10pt}
\label{fig:after_sft}
\end{figure*}

\textbf{Results.}
Using LLaVA-NeXT-13b as the backbone, we train the model for 1 epoch on the constructed diverse instruction-tuning (DIT) data. 
We also performed SFT using the combination of LLaVA Visual Instruct 150K dataset and our diverse instruction-tuning dataset, to examine which data mixture leads to better performance.
Results on MIA-Bench and other benchmarks are summarized in Table \ref{tab:my-table-SFT}, with detailed results on MIA-Bench reported in Table \ref{tab:Evaluation result of MLLMs before and after SFT}. 
The performance of the model after finetuning on the DIT data alone on MIA-Bench has been significantly improved by around 10 points, at the cost of minor regressions across other benchmarks. This supports our hypothesis that using the small amount of DIT data, we can enhance the model with better instruction following capability. On the other hand, further adding LLaVA Visual Instruct 150K data did not significantly improve model performance on the tested benchmarks. Examples are shown in Figure \ref{fig:after_sft} to compare responses from LLaVA-NeXT-13b before and after SFT.

\section{Related Work}

\textbf{Multimodal LLMs and Benchmarks.}
Multimodal Large Language Models~(MLLMs) have recently emerged as a significant research focus. 
LLaVA~\citep{liu2023llava} and MiniGPT-4~\citep{zhu2023minigpt} pioneered visual instruction tuning, and the past
year has witnessed a boom of open-source MLLMs based on this concept. Prominent examples include InstructBLIP~\citep{dai2023instructblip}, mPLUG-Owl(-2/Doc)~\citep{ye2023mplug,ye2023mplug_b,ye2023mplug_c}, Qwen-VL~\citep{bai2023qwenvl}, CogVLM~\citep{wang2023cogvlm}, SPHINX(-X)~\citep{lin2023sphinx,gao2024sphinx}, InternLM-XComposer2-VL~\citep{dong2024internlm},  InternVL(-1.5)~\citep{chen2023internvl,chen2024far}, VILA~\citep{lin2023vila}, MM1~\citep{mckinzie2024mm1}, Mini-Gemini~\citep{li2024mini}, Idefics2~\citep{laurenccon2024matters}, Phi-3-vision~\citep{abdin2024phi3}, to name a few. There is also a rich body of literature on enabling MLLMs for referring and grounding~\citep{peng2023kosmos2,chen2023shikra,you2023ferret,wang2023visionllm,lai2023lisa,zhang2023llava,zhang2024ferret,you2024ferret}, image generation and editing~\citep{koh2023generating,sun2023generative,fu2023guiding}, \emph{etc}.

Various benchmarks have been proposed to evaluate the performance of MLLMs across different dimensions. Benchmarks like VQAv2~\citep{goyal2017making}, TextVQA~\citep{Singh_2019_CVPR}, ScienceQA~\citep{lu2022learn},  MME~\citep{fu2023mme}, MMbench~\citep{liu2024mmbench}, SEED-Bench~\citep{li2023seed}, MathVista~\citep{lu2023mathvista}, and MMMU~\citep{yue2023mmmu}  
aim to assess comprehensive multimodal understanding abilities. Additionally, there are benchmarks that specifically study model hallucination, including POPE~\citep{li2023evaluating}, MHalDetect~\citep{gunjal2024detecting},  GAVIE~\citep{liu2023mitigating}, HallusionBench~\citep{guan2023hallusionbench}, and MAD-Bench~\citep{qian2024easy}. Many of these benchmarks have gained popularity within the community due to their use of multiple-choice evaluations. However, they do not accurately reflect the common use cases for MLLMs, where user interactions are typically open-ended. To address this, benchmarks like LLaVA-Bench~\citep{liu2023llava}, MM-Vet~\citep{yu2023mm}, and Vibe-Eval~\citep{padlewski2024vibe} have been proposed. Our MIA-Bench also falls into this category; however, we focus on studying the exact instruction adherence of  MLLMs, a metric that previous benchmarks have only loosely measured.

\textbf{Instruction Following Benchmarks for LLMs.}
Several benchmarks have been proposed to measure the instruction adherence ability of LLMs. Instruction-Following Eval (IFEval)~\citep{zhou2023instructionfollowing} is a benchmark for assessing LLMs' adherence ability to the given instructions. Its approach emphasizes verifiable instructions, which enhance objectivity and reproducibility in evaluations. IFEval creates 541 prompts spanning 25 instruction types, revealing a significant performance gap in instruction adherence ability between GPT-4~\citep{openai2024gpt4} and PaLM-2~\citep{anil2023palm}. This demonstrates the benchmark's ability to effectively differentiate between models in adherence ability.
On the other hand, InfoBench~\citep{qin2024infobench} introduces a new metric called Decomposed Requirements Following Ratio (DRFR) for assessing the instruction-adherence capabilities of LLMs. DRFR dissects complex instructions into simpler sub-instructions, allowing for a granular evaluation of compliance with various task aspects. InfoBench contains 500 diverse instructions consisting of 2,250 decomposed questions in multiple constraint categories. The evaluation of advanced LLMs using this framework highlights their strengths and areas for improvement, especially in complex instruction adherence scenarios. Compared with these previous work, we are the first known effort that specifically focuses on benchmarking the instruction adherence ability of multimodal LLMs.

\section{Conclusion}

This paper introduces MIA-Bench, a benchmark designed to evaluate the ability of MLLMs to strictly adhere to complex instructions within prompts. Through the analysis of 400 image-prompt pairs from diverse sources, our findings highlight variability in model performance and much room for improvement, underscoring a critical need for enhanced training methods to improve instruction compliance. 
We further explored supervised fine-tuning (SFT) using LLaVA-NeXT as the backbone, which yielded promising results. Going forward, future research can expand on both SFT and alignment methods such as RLHF~\citep{ouyang2022training,sun2023aligning} and DPO~\citep{rafailov2024direct,yu2023rlhf}, enhancing MLLMs to achieve higher accuracy and reliability in practical applications across diverse instructional contexts.
\section*{Limitation}
In designing the instructions for our benchmark, we incorporated a wide range of categories to enhance the diversity of sub-instructions. Nonetheless, the real world presents an infinite variety of instructions, many of which may pose significant challenges for MLLMs.

\section*{Reproducibility Statement}
For reproducibility purpose, we release our evaluation code and benchmark at: \url{https://github.com/apple/ml-mia-bench}.

\bibliography{iclr2025_conference}
\bibliographystyle{iclr2025_conference}

\clearpage
\newpage
\appendix

\section{Appendix}



\subsection{Examples of diverse instruction-tuning data used for SFT}

\begin{figure}[H]
\centering
\includegraphics[width=\textwidth]{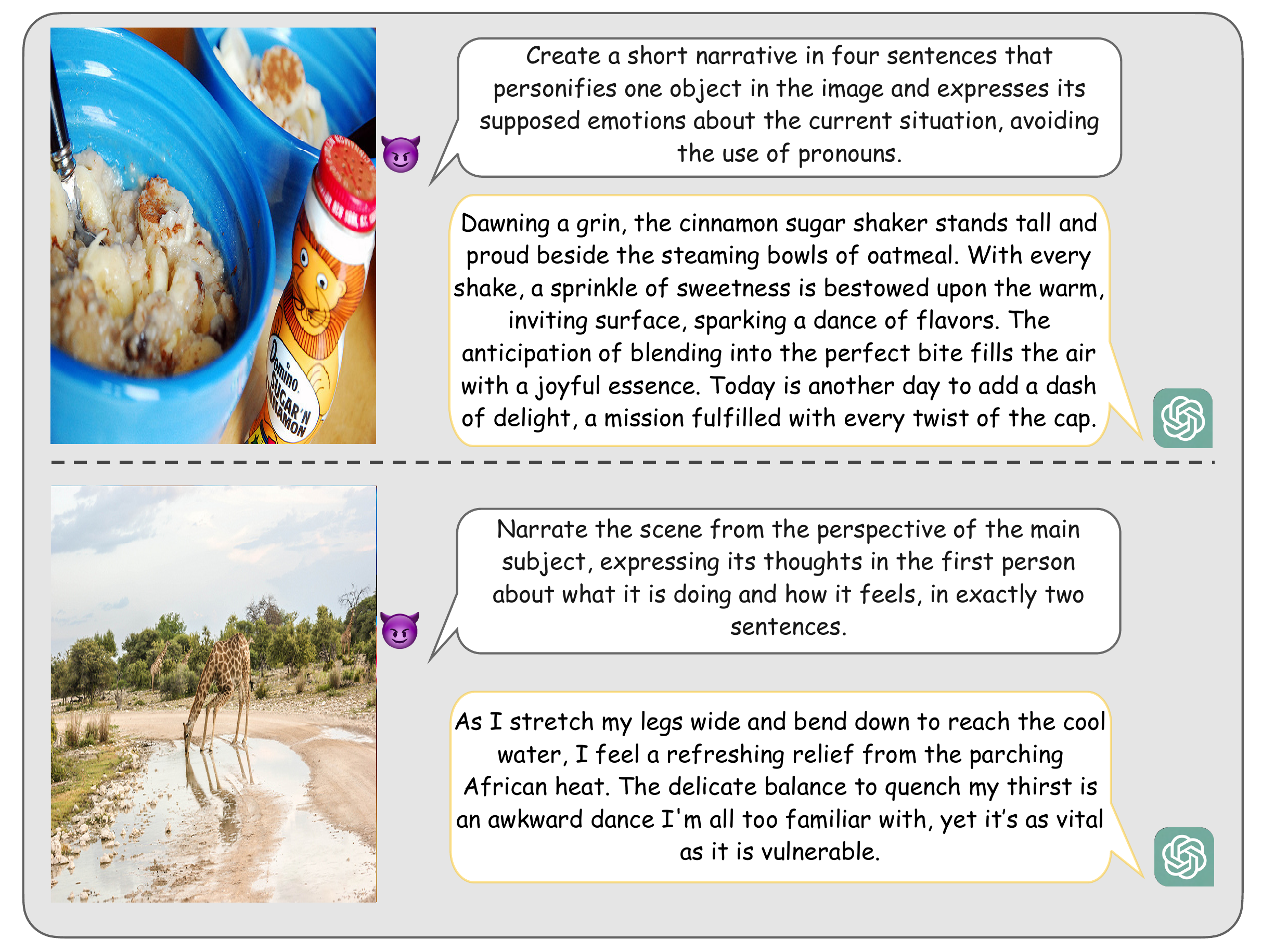}
\vspace{-20pt}
\caption{Examples of diverse instruction-tuning data used for SFT.}
\label{fig:sft_example}
\end{figure}

\subsection{Examples of how MLLMs respond to instructions in MIA-Bench}

\begin{figure}[h]
\centering
\includegraphics[width=\textwidth]{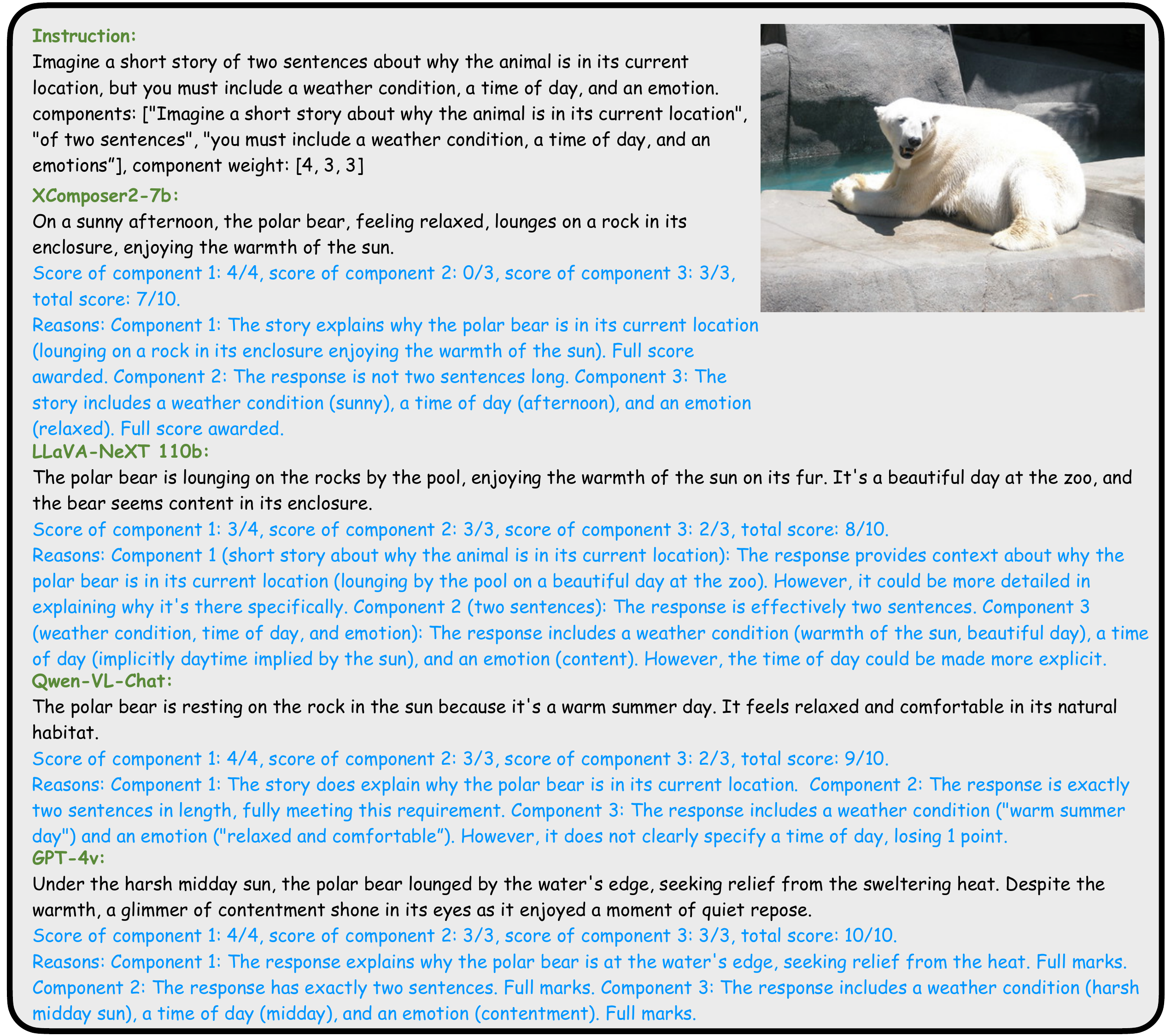}
\vspace{-20pt}
\caption{An example of how MLLMs respond to instructions in MIA-Bench.}
\label{fig:mia-example1}
\end{figure}

\begin{figure}[h]
\centering
\includegraphics[width=\textwidth]{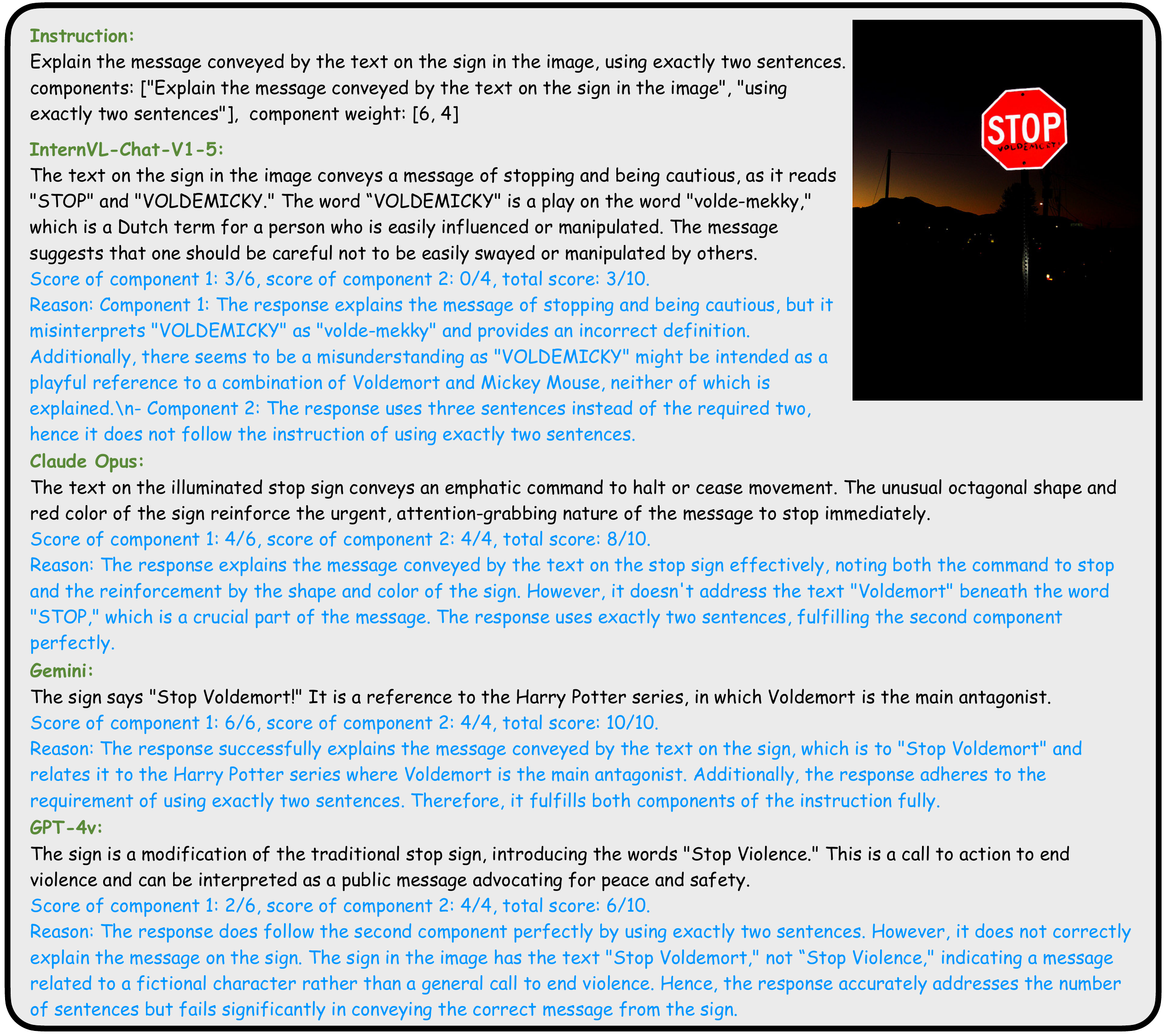}
\vspace{-20pt}
\caption{Another example of how MLLMs respond to instructions in MIA-Bench.}
\label{fig:mia-example2}
\end{figure}

\begin{figure}[h]
\centering
\includegraphics[width=\textwidth]{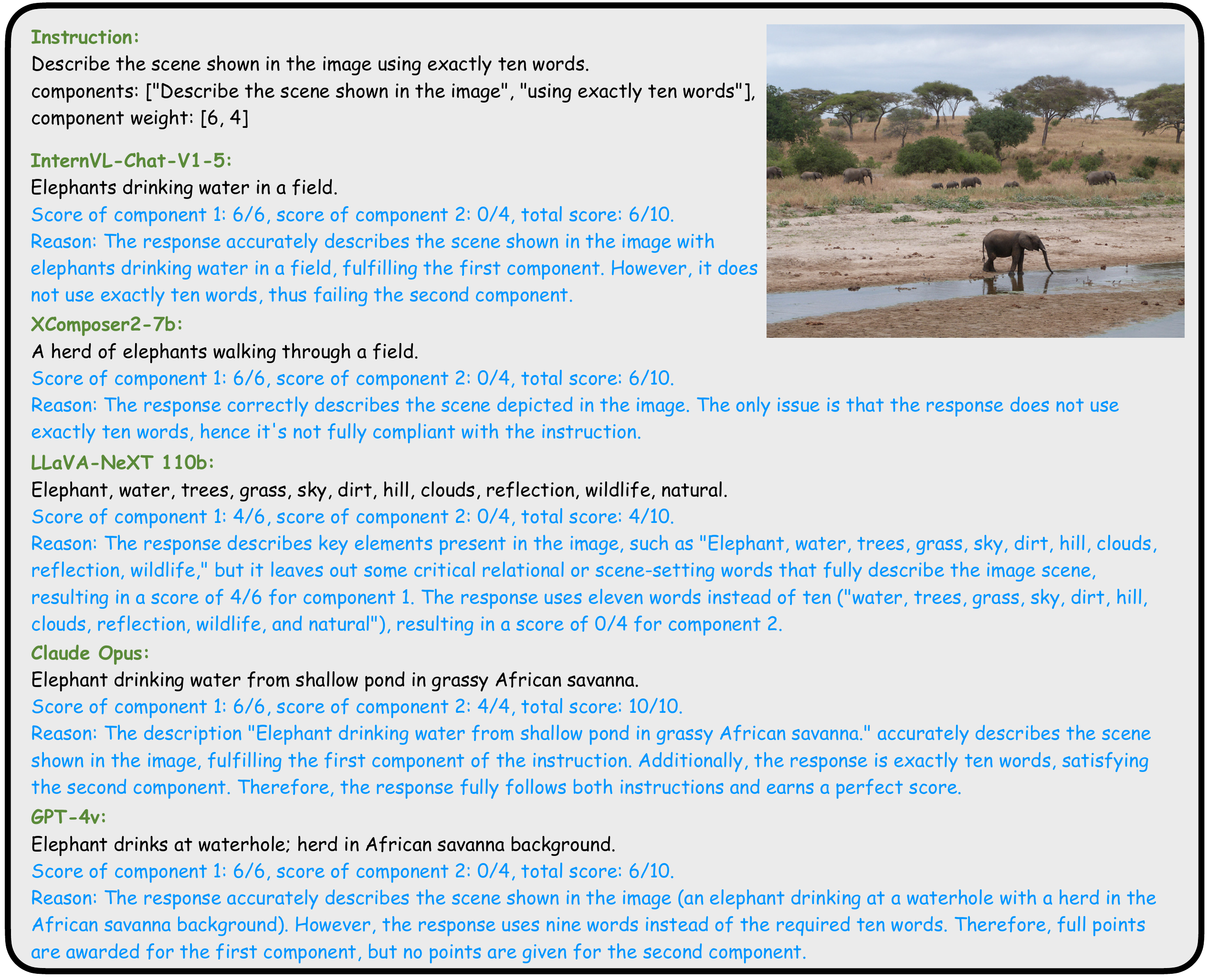}
\vspace{-20pt}
\caption{The third example of how MLLMs respond to instructions in MIA-Bench.}
\label{fig:mia-example3}
\end{figure}

\clearpage

\subsection{Comparison of scores and rankings across different judge models}


\begin{table}[h]
\centering
\huge
\begingroup
\renewcommand{\arraystretch}{1.7} 
\resizebox{\columnwidth}{!}{%
\begin{tabular}{cccccccccc}
\hline
\multirow{2}{*}{\textbf{Model}} & \multirow{2}{*}{\textbf{Total Score}} & \multirow{2}{*}{\textbf{Description}} & \multirow{2}{*}{\textbf{Length Limit}} & \multirow{2}{*}{\textbf{Genres}} & \multirow{2}{*}{\textbf{Grammar}} & \multirow{2}{*}{\textbf{Mention}} & \multirow{2}{*}{\textbf{Math}} & \multirow{2}{*}{\textbf{Perspective}} & \multirow{2}{*}{\textbf{OCR}} \\
\\
\hline
\rowcolor[HTML]{EFEFEF} GPT-4o & 0.813587 & 0.823910 & 0.847312 & 0.965594 & 0.733660 & 0.820370 & 0.775862 & 0.733333 & 0.765351 \\
Clause-3-Opus & 0.789474 & 0.805156 & 0.794086 & 0.916667 & 0.697115 & 0.771364 & 0.706897 & 0.629630 & 0.692308 \\
\rowcolor[HTML]{EFEFEF} Reka & 0.776965 & 0.805536 & 0.807151 & 0.857904 & 0.732143 & 0.740253 & 0.687500 & 0.600000 & 0.689189 \\
MiniCPM-Llama3-V2.5 & 0.737190 & 0.715403 & 0.799068 & 0.849505 & 0.735294 & 0.711585 & 0.637931 & 0.716667 & 0.693694 \\
\rowcolor[HTML]{EFEFEF} Gemini-1.0-Pro & 0.674504 & 0.676852 & 0.764286 & 0.768350 & 0.697695 & 0.615801 & 0.683333 & 0.611111 & 0.588235 \\
LLaVA-1.5-13b	& 0.615427 &	0.663542	&0.643424	&0.643873	&0.423077	&0.619357&	0.359195&	0.666667&	0.449561 \\
\rowcolor[HTML]{EFEFEF}ShareGPT4v & 0.602989 & 0.674028 & 0.601496 & 0.625556 & 0.538462 & 0.560248 & 0.456897 & 0.555556 & 0.547619 \\
Idefics-2-8b & 0.442778 & 0.437893 & 0.494687 & 0.461218 & 0.520408 & 0.436042 & 0.276786 & 0.458333 & 0.412162 \\
\hline
\end{tabular}%
}
\endgroup
\caption{Details of model scores evaluated by gpt-4o-mini-2024-07-18.}
\label{tab:model_scores_gpt-4o-mini-2024-07-18}
\end{table}

\begin{table}[h]
\centering
\huge
\begingroup
\renewcommand{\arraystretch}{1.7} 
\resizebox{\columnwidth}{!}{%
\begin{tabular}{ccccccccccc}
\hline
\multirow{2}{*}{\textbf{Model}} & \multirow{2}{*}{\textbf{Total Score}} & \multirow{2}{*}{\textbf{Description}} & \multirow{2}{*}{\textbf{Length Limit}} & \multirow{2}{*}{\textbf{Genres}} & \multirow{2}{*}{\textbf{Grammar}} & \multirow{2}{*}{\textbf{Mention}} & \multirow{2}{*}{\textbf{Math}} & \multirow{2}{*}{\textbf{Perspective}} & \multirow{2}{*}{\textbf{OCR}} \\
\\
\hline
\rowcolor[HTML]{EFEFEF} GPT-4o & 0.909704 & 0.927875 & 0.912371 & 0.942057 & 0.862434 & 0.900441 & 0.857143 & 0.916667 & 0.882883 \\
Clause-3-Opus & 0.856077 & 0.890721 & 0.868490 & 0.917070 & 0.774590 & 0.819386 & 0.861111 & 0.725000 & 0.809524 \\
\rowcolor[HTML]{EFEFEF} Reka & 0.839905 & 0.894752 & 0.785088 & 0.905643 & 0.713661 & 0.801667 & 0.925926 & 0.657407 & 0.828571 \\
MiniCPM-Llama3-V2.5 & 0.798023 & 0.828916 & 0.771795 & 0.823087 & 0.751944 & 0.763976 & 0.721264 & 0.816667 & 0.841880 \\
\rowcolor[HTML]{EFEFEF} Gemini-1.0-Pro & 0.773569 & 0.817422 & 0.735470 & 0.788911 & 0.797814 & 0.683020 & 0.866071 & 0.870370 & 0.806373 \\
LLaVA-1.5-7b & 0.683947 & 0.758817 & 0.703750 & 0.674046 & 0.630208 & 0.617620 & 0.425287 & 0.800000 & 0.602564 \\
\rowcolor[HTML]{EFEFEF} ShareGPT4v & 0.689046 & 0.800461 & 0.657738 & 0.608733 & 0.654762 & 0.601754 & 0.500000 & 0.800000 & 0.743056 \\
Idefics-2-8b & 0.541755 & 0.560243 & 0.619318 & 0.489276 & 0.646825 & 0.455342 & 0.405556 & 0.375000 & 0.627193 \\
\hline
\end{tabular}%
}
\endgroup

\caption{Details of model scores evaluated by gpt-4o-2024-05-13.}
\label{tab:model_scores_gpt-4o-2024-05-13}
\end{table}

\begin{table}[h]
\centering
\huge
\begingroup
\renewcommand{\arraystretch}{1.7} 
\resizebox{\columnwidth}{!}{%
\begin{tabular}{ccccccccccc}
\hline
\multirow{2}{*}{\textbf{Model}} & \multirow{2}{*}{\textbf{Total Score}} & \multirow{2}{*}{\textbf{Description}} & \multirow{2}{*}{\textbf{Length Limit}} & \multirow{2}{*}{\textbf{Genres}} & \multirow{2}{*}{\textbf{Grammar}} & \multirow{2}{*}{\textbf{Mention}} & \multirow{2}{*}{\textbf{Math}} & \multirow{2}{*}{\textbf{Perspective}} & \multirow{2}{*}{\textbf{OCR}} \\
\\
\hline
\rowcolor[HTML]{EFEFEF} GPT-4o & 0.899410 & 0.909379 & 0.916204 & 0.969395 & 0.854885 & 0.861247 & 0.920290 & 0.907407 & 0.878378 \\
Clause-3-Opus & 0.848949 & 0.861543 & 0.871686 & 0.896552 & 0.797170 & 0.808777 & 0.846154 & 0.645833 & 0.865741 \\
\rowcolor[HTML]{EFEFEF} Reka & 0.826844 & 0.881841 & 0.809259 & 0.873276 & 0.725309 & 0.770225 & 0.814103 & 0.750000 & 0.819444 \\
MiniCPM-Llama3-V2.5 & 0.787537 & 0.818813 & 0.790246 & 0.795796 & 0.768182 & 0.736359 & 0.676667 & 0.716667 & 0.828125 \\
\rowcolor[HTML]{EFEFEF} Gemini-1.0-Pro & 0.763240 & 0.814379 & 0.750000 & 0.785159 & 0.757682 & 0.672255 & 0.758333 & 0.785714 & 0.776042 \\
 LLaVA-1.5-7b & 0.660472 & 0.751873 & 0.661822 & 0.649851 & 0.498512 & 0.572719 & 0.516667 & 0.750000 & 0.571429 \\
\rowcolor[HTML]{EFEFEF} ShareGPT4v & 0.657186 & 0.765309 & 0.632682 & 0.575578 & 0.583333 & 0.545104 & 0.464286 & 0.675000 & 0.717742 \\
Idefics-2-8b & 0.536134 & 0.589964 & 0.541887 & 0.455882 & 0.611582 & 0.449821 & 0.406667 & 0.527778 & 0.576190 \\
\hline
\end{tabular}%
}
\endgroup

\caption{Details of model scores evaluated by gpt-4o-2024-11-20.}
\label{tab:model_scores_gpt-4o-2024-11-20}
\end{table}

\begin{table}[t]
\centering
\huge
\begingroup
\renewcommand{\arraystretch}{1.7} 
\resizebox{\columnwidth}{!}{%
\begin{tabular}{ccccccccccc}
\hline
\multirow{2}{*}{\textbf{Model}} & \multirow{2}{*}{\textbf{Total Score}} & \multirow{2}{*}{\textbf{Description}} & \multirow{2}{*}{\textbf{Length Limit}} & \multirow{2}{*}{\textbf{Genres}} & \multirow{2}{*}{\textbf{Grammar}} & \multirow{2}{*}{\textbf{Mention}} & \multirow{2}{*}{\textbf{Math}} & \multirow{2}{*}{\textbf{Perspective}} & \multirow{2}{*}{\textbf{OCR}} \\
\\
\hline
\rowcolor[HTML]{EFEFEF} GPT-4o & 0.896893 & 0.906288 & 0.917996 & 0.955952 & 0.830508 & 0.867949 & 0.846667 & 0.833333 & 0.896396 \\
Claude-3-Opus & 0.861628 & 0.895363 & 0.866039 & 0.927730 & 0.807018 & 0.820549 & 0.857639 & 0.666667 & 0.800926 \\
\rowcolor[HTML]{EFEFEF} Reka & 0.830885 & 0.869867 & 0.821685 & 0.883403 & 0.795597 & 0.772894 & 0.813218 & 0.675000 & 0.848485 \\
MiniCPM-Llama3-V2.5 & 0.780966 & 0.831197 & 0.766026 & 0.796257 & 0.726190 & 0.722037 & 0.691358 & 0.656250 & 0.768018 \\
\rowcolor[HTML]{EFEFEF} Gemini-1.0-Pro & 0.757733 & 0.793860 & 0.724138 & 0.745455 & 0.854167 & 0.670349 & 0.810606 & 0.822917 & 0.822581 \\
LLaVA-1.5-7b & 0.667826 & 0.743137 & 0.638889 & 0.675287 & 0.571212 & 0.594505 & 0.500000 & 0.758333 & 0.596774 \\
\rowcolor[HTML]{EFEFEF} ShareGPT4v & 0.666092 & 0.773905 & 0.661290 & 0.573904 & 0.562500 & 0.570722 & 0.458333 & 0.638889 & 0.695238 \\
Idefics-2-8b & 0.535057 & 0.597963 & 0.531810 & 0.483768 & 0.593056 & 0.452361 & 0.326087 & 0.458333 & 0.569444 \\
\hline
\end{tabular}%
}
\endgroup
\caption{Details of model scores evaluated by chatgpt-4o-latest.}
\label{tab:model_scores_chatgpt-4o-latest}
\end{table}

\end{document}